\documentclass{article}
\usepackage{spconf,amsmath,graphicx}

\usepackage{microtype}
\usepackage{amssymb}
\usepackage{subfigure}
\usepackage{threeparttable}
\usepackage{multirow}
\usepackage{booktabs}

\usepackage{color}

\title{Improving the Out-Of-Distribution Generalization Capability of Language Models: Counterfactually-Augmented Data is not Enough}

\name{Caoyun Fan$^{1 \dagger}$ \quad Wenqing Chen$^{2 \dagger}$ \quad Jidong Tian$^{1}$ \quad Yitian Li$^{1}$ \quad Hao He$^{1 *}$ \quad Yaohui Jin$^{1 *}$}
\address{$^{1}$ MoE Key Lab of Artificial Intelligence, AI Institute, Shanghai Jiao Tong University, China \\ $^{2}$ School of Software Engineering, Sun Yat-sen University, China}

\begin{document}
%\ninept
%
\maketitle
\begin{abstract}

% Numerous studies have shown that models have poor out-of-distribution (OOD) generalization capability in Natural Language Processing due to their reliance on spurious correlations. 
Counterfactually-Augmented Data (CAD) has the potential to improve language models' Out-Of-Distribution (OOD) generalization capability, as CAD induces language models to exploit causal features and exclude spurious correlations. However, the empirical results of OOD generalization on CAD are not as efficient as expected. In this paper, we attribute the inefficiency to \textbf{Myopia Phenomenon} caused by CAD: language models only focus on causal features that are edited in the augmentation and exclude other non-edited causal features. As a result, the potential of CAD is not fully exploited. Based on the structural properties of CAD, we design two additional constraints to help language models extract more complete causal features contained in CAD, thus improving the OOD generalization capability. We evaluate our method on two tasks: Sentiment Analysis and Natural Language Inference, and the experimental results demonstrate that our method could unlock CAD's potential and improve language models' OOD generalization capability. % \uncertain{ok}

% Counterfactually-Augmented Data (CAD) -- minimal editing of sentences to flip the corresponding labels -- has the potential to improve language models' Out-Of-Distribution (OOD) generalization capability, as CAD induces models to exploit domain-independent causal features and exclude the spurious correlations interference. However, the empirical results of OOD generalization using CAD are not as efficient as expected. In this paper, we attribute the inefficiency to \textbf{Myopia Phenomenon} caused by CAD: models only focus on causal features that are edited in the augmentation and ignore other non-edited causal features. As a result, the potential of CAD is not fully exploited. Based on CAD's structural properties of CAD (dataset-level and sentence-level), we design two additional constraints to help models extract more complete causal features contained in CAD, thus improving the models' OOD generalization capability. We carefully evaluate our method on two tasks: Sentiment Analysis and Natural Language Inference, and the experimental results show that our method could unlock the potential of CAD and improve language models' OOD generalization capability. % \uncertain{ok}

\end{abstract}

\section{Introduction}

\renewcommand{\thefootnote}{\fnsymbol{footnote}}
\footnotetext[1]{Corresponding author. }
\footnotetext[2]{These authors contributed equally. }
\renewcommand{\thefootnote}{\arabic{footnote}}

Despite the remarkable performance of language models in Natural Language Processing (NLP) \cite{Devlin2019BERTPO,Liu2019RoBERTaAR}, the Out-Of-Distribution (OOD) generalization capability of language models is often disappointing \cite{Wang2021GeneralizingTU,Shen2021TowardsOG}. Many studies \cite{Teney2020LearningWM,Joshi2021AnIO} have pointed out that such limited generalization capability partly arises from the language models' exploitation of spurious correlations \cite{Mitchell2007TheNF,Torralba2011UnbiasedLA,McCoy2019RightFT,Wang2020IdentifyingSC} in the dataset. Specifically, the language models tend to exploit dataset-specific correlation bias \cite{Teney2020LearningWM,Spears2004EvaluationAS} rather than the intrinsic properties of tasks to make predictions during the training process, while the spurious correlations can not be generalized to OOD data. 

% transport out of the domain. 

% \begin{figure}[t]
%     \centering
%     \includegraphics[width=1.0\columnwidth]{fig/CAD example.pdf}
%     % \vspace{-0.6cm}
%     \caption{An example of counterfactual sentence pairs from in Sentiment Analysis task. The motivation of CAD is that the model exploits the causal features (in gray) in the sentence and ignores the possible spurious correlations (e.g., \emph{Nolan} in the sentence). }
%     \label{f1-1}
%     % \vspace{-0.3cm}
% \end{figure}

\begin{figure}
    \centering
    \subfigure[An example of counterfactual sentence pairs. We expect the language model to exploit the causal features (in gray) and exclude the possible spurious correlations (e.g., \emph{Nolan} in the sentence). ]{
    \includegraphics[width=0.9\columnwidth]{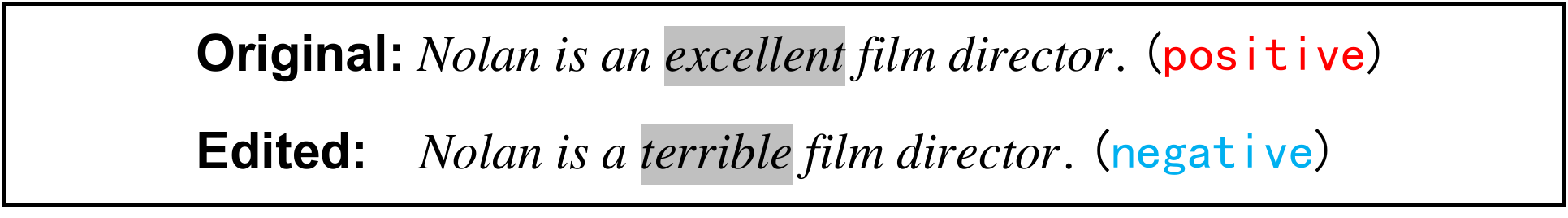}
    \label{f1-1a}
    }
    \subfigure[Original Dataset]{
    \includegraphics[height=0.35\columnwidth]{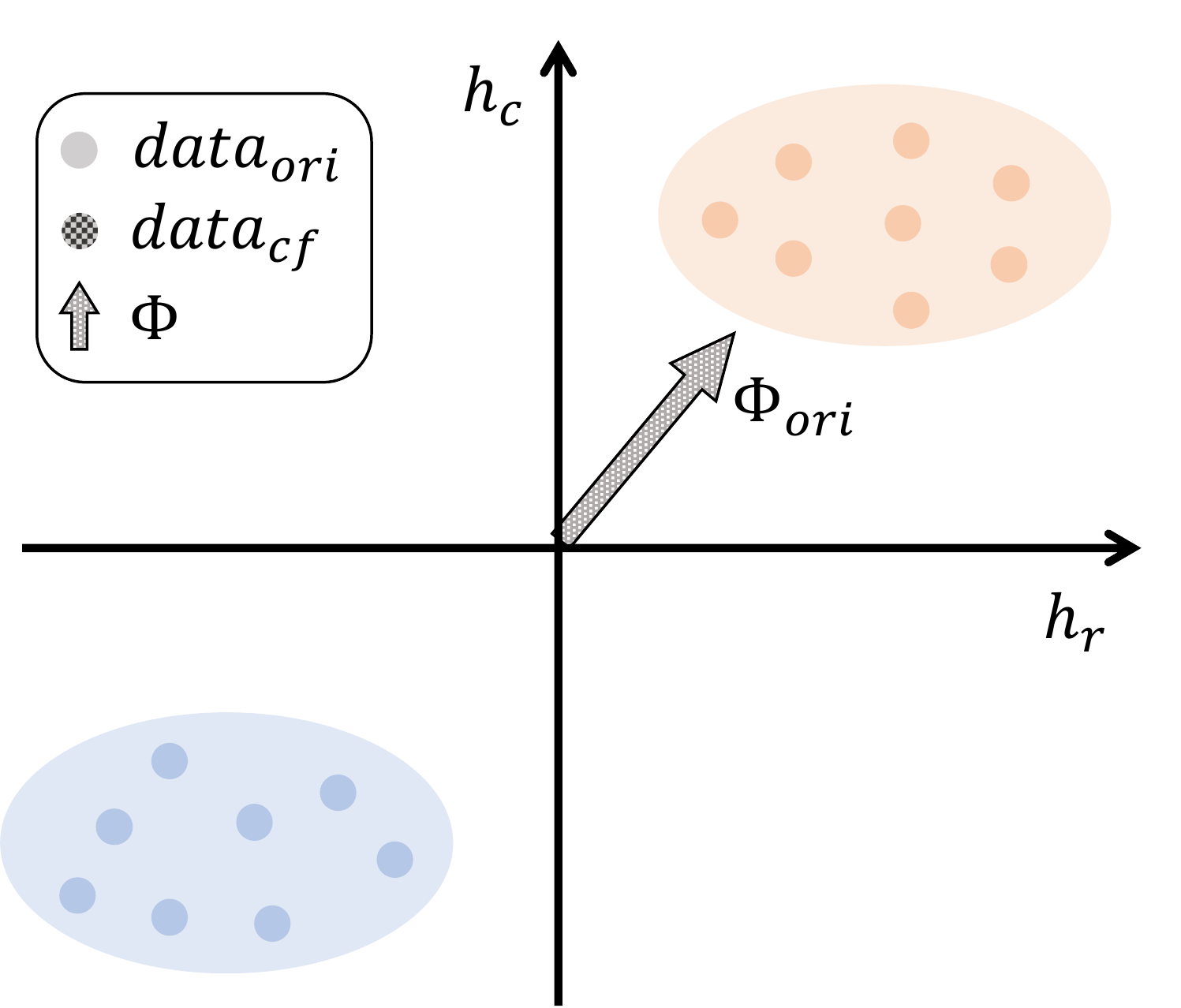} 
    \label{f1-1b}
    }
    \subfigure[CAD]{
    \includegraphics[height=0.35\columnwidth]{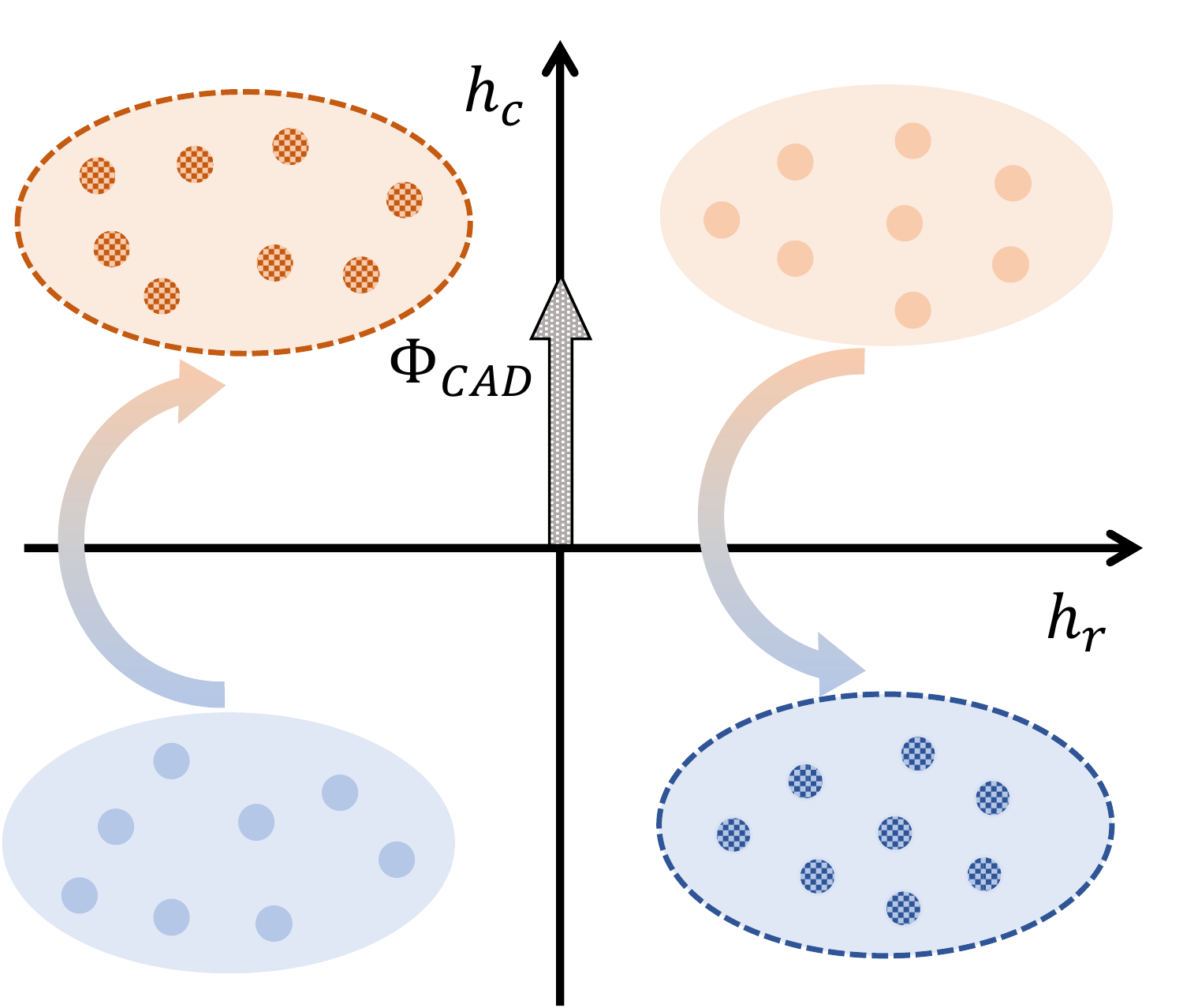} 
    \label{f1-1c}
    }
    \caption{The motivation of CAD. Counterfactual augmentation of texts (Fig. \ref{f1-1a}) changes the data distribution of the dataset (from Fig. \ref{f1-1b} to Fig. \ref{f1-1c}), which helps the model $\Phi$ to exploit causal features $h_c$ and exclude correlated features $h_r$. }
    \label{f1-1}
\end{figure}

% \noindent\footnotetext[1]{Throughout the rest of the paper, the counterfactual sentence pair refers to the pair consisting of the original sentence and the corresponding edited sentence. }

% The motivation of CAD is that the model could distinguish between spurious correlations and intrinsic properties by comparing counterfactual sentence pairs. 

To solve the problem of spurious correlations, a recent promising direction is Counterfactually-Augmented Data (CAD) \cite{Kaushik2020LearningTD,Wang2021RobustnessTS,Yang2021ExploringTE}: minimal editing of sentence to flip the corresponding label $Y$, where the edited part is considered to be the intrinsic properties of the task and have a causal effect on the label (Fig. \ref{f1-1a}). Unlike the Independent Identically Distribution (IID) principle of most data augmentation methods, CAD aims to change the data distribution of the dataset so that the language models can alleviate reliance on dataset-specific bias and exclude spurious correlations. 

Under the ideal conditions assumed by \cite{Joshi2021AnIO} \footnote{Under the ideal conditions, each sentence consists of causal features $h_c$ whose joint distribution with labels is invariant, and correlated features $h_r$ whose joint distribution can vary. }, CAD keeps correlated features $h_r$ in the counterfactual sentence pairs constant while the causal features $h_c$ change. Therefore, the classifier $\Phi$ can make predictions based on causal features and then exclude the interference of spurious correlations as: 
% \vspace{-0.1cm}
\begin{equation}
    \begin{split}
        \Phi(h_c, h_r) &= Y \\
        \Phi(h_c^*, h_r) &= Y^*
    \end{split}
    \label{e1-1}
\end{equation}

% Intuitively, there are more causal components in complex tasks. 
\begin{figure*}[t]
    \centering
    % \vspace{-0.7cm}
    \includegraphics[width=1.55\columnwidth]{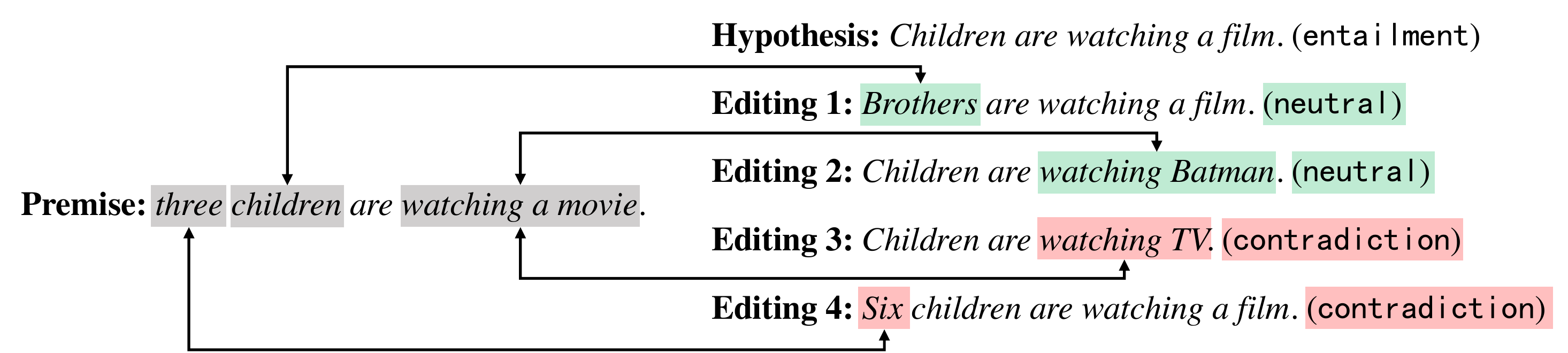}
    \caption{An example of multiple counterfactual augmentation results in Natural Language Inference. Editing different causal components in the \textbf{Hypothesis} can all serve the purpose of flipping the corresponding label. }
    \label{f2-1}
    % \vspace{-0.5cm}
\end{figure*}

\noindent where $h_c^*$ and $Y^*$ are the causal features and the label of the edited sentence, respectively. Intuitively, the classifier $\Phi$ no longer focuses on $h_r$ because different labels correspond to the same $h_r$, as shown in Fig. \ref{f1-1b} \& \ref{f1-1c}. However, some experiments \cite{Huang2020CounterfactuallyAugmentedST,Khashabi2020MoreBF} have demonstrated that CAD is not efficient in improving the generalization capability of language models, especially in more complex tasks. This is not in line with our expectations for CAD. 

In this work, we attribute the inefficiency of CAD in generalization to the CAD-imposed myopia phenomenon: language models focus only on causal features edited by counterfactual augmentation, which means correlated features along with other non-edited causal features are excluded. However, all causal features are beneficial for OOD generalization \cite{Balashankar2021CanWI}. To \textbf{E}xtract more complete \textbf{C}ausal \textbf{F}eatures and unlock the potential of CAD for language models, we design the ECF algorithm: introducing additional constraints in the training process based on the structural properties of CAD. Specifically, we extract invariant causal features over both distributions of CAD and the original dataset by the Invariant Risk Minimization \cite{Rosenfeld2021TheRO} method (dataset level) and constrain the correlated feature similarity of counterfactual sentence pairs (sentence level). Through extensive experiments across multiple language models and NLP tasks, we conclude that the proposed ECF algorithm could help language models to extract more complete causal features, and then improve the OOD generalization capability in multiple NLP tasks. % \uncertain{ok}

% Most traditional machine learning models in NLP are based on the i.i.d. the assumption that training and testing data are identically and independently distributed \cite{Vapnik1999AnOO}, and the models identify correlations between given sentences and their labels in the training set and make predictions in the testing set. However, the models obtained in this paradigm face an unavoidable challenge: there are spurious correlations (e.g., dataset-specific bias and sampling artifacts \cite{Spears2004EvaluationAS,Teney2020LearningWM}) in the training set, and the models lack the capability to discriminate between the intrinsic properties of the task and the spurious correlations \cite{Mitchell2007TheNF,Torralba2011UnbiasedLA}. Therefore, such models’ predictive performance suffers and the outputs become unreliable and unpredictable on OOD data because the spurious correlations no longer hold. 

% Numerous studies \cite{Teney2020LearningWM,Wang2021CounterfactualAL} have pointed out that the key to improving the stability and generalization of models on OOD data is to capture the causal mechanisms behind the data while excluding the spurious and correlation interference. A recent promising direction is the collection of counterfactual enhanced data (CAD). 

% \vspace{-0.1cm}

\section{Myopia Phenomenon in CAD}
\label{s2}

% \footnote{\cite{Joshi2021AnIO} provided a detailed summary of the perturbation types in CAD}
As mentioned before, the essence of CAD is to change the data distribution through data augmentation, thereby reducing the dataset-specific bias implied in the data distribution. Intuitively, by comparing the differences in counterfactual sentence pairs, language models could capture the features that have a causal effect on the labels. However, the results of counterfactual augmentation are diverse for a particular sentence, as illustrated in Fig. \ref{f2-1}. Specifically, the causal components and the perturbation types \cite{Joshi2021AnIO} (e.g., negation, quantifier, lexical, delete) that can flip labels are diverse, so the different counterfactual sentence can be obtained by making a specific perturbation for a particular causal component, while the other causal components remain unchanged. 

Therefore, compared to Eq. \ref{e1-1}, a more reasonable assumption is that only part of $h_c$ in the counterfactual sentence pairs would change with the counterfactual augmentation as: 
% \vspace{-0.1cm}
\begin{equation}
    \begin{split}
        \Phi(h_e, h_u, h_r) &= Y \\
        \Phi(h_e^*, h_u, h_r) &= Y^*
    \end{split}
    \label{e2-1}
\end{equation}

\noindent where $h_c$ is distinguished into edited features $h_e$ that change with augmentation and non-edited features $h_u$ that do not change. This assumption is empirically convincing because of the analysis and experiments in \cite{Joshi2021AnIO,Kaushik2020LearningTD}. Similar to the analysis of Eq. \ref{e1-1}, Eq. \ref{e2-1} gives us an important insight: language models trained on original data and CAD focus on different features in the sentence. On the one hand, CAD eliminates the interference of correlated features; on the other hand, language models inevitably ignore non-edited causal features. In this paper, we refer to this as \textbf{Myopia Phenomenon}. % \uncertain{ok}

% \vspace{-0.1cm}

\section{Proposed Method}
\label{s3}

% mentioned in Section \ref{s2}
To solve the myopia phenomenon and extract more complete causal features, we propose two insights on the structural properties of CAD at the dataset level and at the sentence level, and design additional constraints based on these insights, to further exploit the generalization potential of CAD. % \uncertain{ok}

\subsection{Dataset-Level Constraint}

\emph{\textbf{Insight:} the data distribution of the original dataset can alleviate the Myopia Phenomenon of CAD. }

% Previous experiments \cite{Kaushik2020LearningTD} have also revealed that models trained on original data performed poorly on corresponding CAD and vice versa. 
Due to the change in data distribution, the features that language models focus on are different: models with CAD only focus on edited causal features $h_e$ (Myopia Phenomenon), while models with the original dataset confuse $h_c$ and $h_r$ (but no Myopia Phenomenon). Different data distributions lead to different problems, which indicates that the original data distribution carries information that is missing in CAD. Therefore, there are potential complementary effects of the original dataset and CAD on causal feature extraction. 

% According to the analysis in Section \ref{s2}, the original data and CAD have different feature distributions, 
% so we consider them as two training environments $\mathcal{E}_{tr} = \{ e_{\text{ori}}, e_{\text{CAD}} \}$, and use IRM to fuse the advantages of both. 
% Although CAD is obtained by extending the original data, this extension is non-identically distributed and can change the feature distribution in the dataset. 
Inspired by \cite{Rosenfeld2021TheRO}, we adopt the Invariant Risk Minimization (IRM) method to extract more complete causal features in CAD. The role of IRM is to estimate invariant causal features from multiple training environments. As mentioned before, counterfactual augmentation does not follow the IID principle, which allows us to consider the original dataset and CAD as two different training environments $\mathcal{E}_{tr} = \{ e_{\text{ori}}, e_{\text{CAD}} \}$, and then adopt the IRM method to fuse the advantages of both environments. Specifically, to induce the language model $M$ to learn invariant causal features across environments, the additional constraint $\mathcal{L}_{IRM}$ is designed as: 
% \vspace{-0.1cm}
\begin{equation}
    \mathcal{L}_{IRM} = \sum_{e \in \mathcal{E}_{tr}} \Vert \nabla_{\omega | \omega=1.0} \mathcal{R}_{e}(\omega \cdot M) \Vert^2
\end{equation}

\begin{table}[t]
  \centering
  \fontsize{10}{10}\selectfont
  \begin{threeparttable}
    \resizebox{8.6cm}{!}{
    \begin{tabular}{llcccccc}
    \toprule
    \bf Dataset & \bf Method & \bf Original & \bf CAD & \bf SST-2 & \bf Amazon & \bf Yelp & \bf Mean \cr
    \midrule
    \multirow{2}{*}{Seed} & LSTM & 74.3 & 64.0 & 64.5 & 61.7 & 62.3 & 62.8 \cr
     & BERT & 85.2 & 85.5 & 82.8 & 88.0 & 85.5 & 85.4 \cr
    \midrule
     & LSTM & 81.0 & 86.7 & 65.3 & \bf 74.0 & 72.7 & 70.7 \cr
    $\text{CAD}_{h}$ & LSTM+ECF & 80.7 & 84.4 & \bf 71.4 & 74.0 & \bf 77.4 & \bf 74.3 \cr
    \cmidrule(lr){2-8}
    \cite{Kaushik2020LearningTD} & BERT & 88.9 & 92.3 & \bf 86.9 & 90.3 & 89.8 & 89.0 \cr
     & BERT+ECF & 84.7 & 89.2 & 84.9 & \bf 92.9 & \bf 92.4 & \bf 90.1 \cr
    \midrule
     & LSTM & 56.1 & 66.7 & 61.5 & 57.6 & 57.9 & 59.0 \cr
    $\text{CAD}_{a}$ & LSTM+ECF & 57.9 & 67.5 & \bf 62.4 & \bf 59.0 & \bf 58.5 & \bf 60.0 \cr
    \cmidrule(lr){2-8}
    \cite{Yang2021ExploringTE} & BERT & 55.1 & 72.1 & 75.9 & 84.7 & 83.1 & 81.2 \cr
     & BERT+ECF & 80.4 & 63.8 & \bf 78.9 & \bf 88.1 & \bf 86.2 & \bf 84.4 \cr
    \midrule
     & LSTM & 67.8 & 75.4 & 59.8 & 63.3 & 62.6 & 61.9 \cr
    $\text{CAD}_{a}$ & LSTM+ECF & 76.1 & 79.7 & \bf 61.7 & \bf 66.4 & \bf 63.8 & \bf 64.0 \cr
    \cmidrule(lr){2-8}
    \cite{Wang2021RobustnessTS} & BERT & 87.1 & 88.0 & 83.2 & 88.4 & 88.9 & 86.8 \cr
     & BERT+ECF & 85.7 & 77.8 & \bf 83.6 & \bf 90.3 & \bf 89.5 & \bf 87.8 \cr
    \bottomrule
    \end{tabular}
    }
    \end{threeparttable}
    % \vspace{-0.1cm}
    \caption{Accuracy of different language models and datasets in SA. The best performance is \textbf{bold}. $\text{CAD}_{h}$ and $\text{CAD}_{a}$ represent manually annotated CAD and automatically generated CAD, respectively. }
    \label{tab5-1}
    % \vspace{-0.5cm}
\end{table}

% To learn invariances across environments, find a data representation such that the optimal classifier on top of that representation matches all environments. 
\noindent where $\mathcal{R}_{e}(\cdot)$ is the risk \cite{Rosenfeld2021TheRO} under environment $e$, and $\omega = 1.0$ as a scalar is a fixed `dummy’ classifier. The essence of $\mathcal{L}_{IRM}$ is a gradient norm penalty that measures the optimality of the `dummy’ classifier in each environment, in order to find invariant causal features that match all environments. % \uncertain{ok}

\subsection{Sentence-Level Constraint}

\emph{\textbf{Insight:} the correlated features $h_r$ of counterfactual sentence pairs are not guaranteed to be aligned. }

% it is consistent with the original design of CAD to make an explicit constraint on $h_r$ for counterfactual sentence pairs. 
In our assumptions, the correlated features $h_r$ of counterfactual sentence pairs are similar, because the augmentation operation only affects part of $h_c$. However, this property is not guaranteed for language models trained directly on CAD, and this potential dissimilarity gives language models the convenience to utilize $h_r$. Therefore, it is reasonable to design an explicit constraint on $h_r$ for counterfactual sentence pairs. 

% most models can be divided into a feature extraction module $\Psi$ and a Fully-Connected layer (concatenated by label vectors $h_Y$
However, $h_r$ and $h_c$ in CAD are hard to decouple in language models, so a sensible proxy for $h_r$ is needed. Noting that $h_r$ has little effect on the prediction in CAD, based on this property, we creatively construct the proxy of $h_r$ using the mechanism of feature classifier. Most feature classifiers are fully-connected layers, where each row of the weight matrix can be interpreted as a label vector $h_Y$ \cite{Du2019ExplicitIM}, and the label probability can be obtained by the dot product of the sentence vector $h$ and each label vector $h_Y$ as: 
\begin{equation}
    p(y_k) = \frac{\text{exp}(h_Y^k \cdot h)}{\sum_{i=1}^N\text{exp}(h_Y^i \cdot h)}
\end{equation}

In this way, $h$ can be decomposed along $h_Y$, where the parallel component $h_{\parallel Y}$ determines the prediction and the orthogonal component $h_{\perp Y}$ has no effect on the prediction. The commonality between $h_{\perp Y}$ and $h_r$ makes $h_{\perp Y}$ an ideal proxy for $h_r$. Specifically, for a counterfactual sentence feature pair $(h, h^*)$, we design $\mathcal{L}_{OCD}$ to penalize their Orthogonal Component Distance as: 
% \vspace{-0.1cm}
\begin{equation}
    \mathcal{L}_{OCD} = \Vert h_{\perp Y} - h^*_{\perp Y^*} \Vert^2
\end{equation}

This is a positive feedback process, so even if initially the classifier has large estimation errors, it will gradually become accurate with the help of the prediction loss and $\mathcal{L}_{OCD}$. % \uncertain{ok}

\subsection{Training Process}

Compared to the original prediction loss $\mathcal{L}_{P}$, the proposed ECF algorithm also combines dataset-level constraint $\mathcal{L}_{IRM}$ and sentence-level constraint $\mathcal{L}_{OCD}$ as: 
% \vspace{-0.1cm}
\begin{equation}
    \mathcal{L} = \mathcal{L}_{P} + \alpha \cdot \mathcal{L}_{IRM} + \beta \cdot \mathcal{L}_{OCD}
\end{equation}

\noindent where $\alpha$, $\beta$ are the weighting coefficients to balance the language model's In-Distribution predictive power and additional constraints introduced for OOD Generalization. % \uncertain{ok}

\begin{table}[t]
  \centering
  \fontsize{10}{10}\selectfont
  \begin{threeparttable}
    \resizebox{8.6cm}{!}{
    \begin{tabular}{llccccc}
    \toprule
    \bf Dataset & \bf Method & \bf Original & \bf CAD & \bf MNLI-m & \bf MNLI-mm & \bf Mean \cr
    \midrule
    \multirow{3}{*}{Seed} & LSTM & 41.8 & 33.9 & 35.9 & 35.0 & 35.5 \cr
     & BERT & 71.5 & 53.8 & 53.6 & 55.1 & 54.3 \cr
     & Roberta & 83.8 & 67.2 & 67.4 & 68.4 & 67.9 \cr
    \midrule
     & LSTM & 39.8 & 39.0 & 34.4 & 35.0 & 34.7 \cr
     & LSTM+ECF & 44.2 & 37.6 & \bf 36.2 & \bf 36.2 & \bf 36.2 \cr
    \cmidrule(lr){2-7}
    $\text{CAD}_{h}$ & BERT & 79.2 & 71.2 & 62.6 & 64.3 & 63.5 \cr
    \cite{Kaushik2020LearningTD} & BERT+ECF & 77.0 & 72.0 & \bf 64.0 & \bf 65.5 & \bf 64.7 \cr
    \cmidrule(lr){2-7}
     & Roberta & 80.2 & 75.4 & 70.5 & 71.5 & 71.0 \cr
     & Roberta+ECF & 82.5 & 76.7 & \bf 72.6 & \bf 72.8 & \bf 72.7 \cr
    \bottomrule
    \end{tabular}
    }
    \end{threeparttable}
    % \vspace{-0.1cm}
    \caption{Accuracy of different language models and datasets in NLI. The best performance is \textbf{bold}. }
    \label{tab5-2}
    % \vspace{-0.3cm}
\end{table}

% \vspace{-0.1cm}

\section{Experiments}

\subsection{Datasets}

We conducted experiments on two tasks: Sentiment Analysis (SA) and Natural Language Inference (NLI). % \uncertain{ok}
% CAD was obtained by extending the seed dataset with counterfactual augmentation. 

\noindent \textbf{Sentiment Analysis} \quad The seed dataset in SA was IMDb \cite{Maas2011LearningWV} dataset. \cite{Kaushik2020LearningTD} collected a subset of IMDb as a seed dataset, and manually annotated the corresponding counterfactual sentences to construct $\text{CAD}_{h}$, while \cite{Yang2021ExploringTE,Wang2021RobustnessTS} utilized WordNet \cite{Fellbaum2000WordNetA} to automatically generate counterfactual sentences and constructed $\text{CAD}_{a}$. We evaluated each language model's OOD generalization capability on three OOD datasets: SST-2 \cite{Socher2013RecursiveDM}, Amazon review \cite{Ni2019JustifyingRU}, Yelp review \cite{Zhang2015CharacterlevelCN}. % \uncertain{ok}

\noindent \textbf{Natural Language Inference} \quad \cite{Kaushik2020LearningTD} constructed $\text{CAD}_{h}$ by manually editing seed dataset from SNLI \cite{Bowman2015ALA} dataset. Because the NLI task is more complex, there is little research related to the automatic generation of counterfactual sentences. We treated MNLI (split into matched and mismatched parts) \cite{Williams2018ABC} as our OOD dataset for evaluation. % \uncertain{ok}

\subsection{Implementation Details}

% Hugging Face implementation \cite{Wolf2019HuggingFacesTS}
We implemented LSTM \cite{Hochreiter1997LongSM} and pre-trained models BERT \cite{Devlin2019BERTPO}, Roberta \cite{Liu2019RoBERTaAR} as our backbones, and selected the best checkpoint on the training set for testing. For LSTM, The word embedding dimension was set to 300, the batch size was set to 32, and the learning rate of the Adam optimizer to 1e-3. We set $\alpha$ = 1.6 and $\beta$ = 0.1. We trained each model for 100 epochs in SA/NLI task. For pre-trained models, we used the Hugging Face implementation to finetune the pre-trained models. The batch size was set to 8/5 for SA/NLI tasks respectively, and the learning rate of Adam optimizer to 1e-5. We set $\alpha$ = 0.1 and $\beta$ = 0.1. We trained each model for 10 epochs. 

% \uncertain{ok}

\subsection{Main Results}

\noindent \textbf{Results on SA} \quad The results are presented in Table \ref{tab5-1}, where the ECF algorithm beat all the compared backbones in terms of the average accuracy of OOD datasets. Specifically, $\text{CAD}_{h}$ was more effective for language models' generalization, while the ECF algorithm improved the average accuracy of LSTM and BERT on OOD datasets by 3.6\% and 1.1\%, respectively. The language models trained on $\text{CAD}_{a}$ were relatively weaker in generalization, and the ECF algorithm also helped LSTM and BERT improve their average accuracy by 1.0\%/2.1\% and 3.2\%/1.0\% on two $\text{CAD}_{a}$, respectively. 

% % \uncertain{?} and it was worth mentioning that the generalization capability of LSTM for $\text{CAD}_{h}$ was even worse than that for the seed dataset. 
\noindent \textbf{Results on NLI} \quad The results are presented in Table \ref{tab5-2}. The ECF algorithm improved the average accuracy of LSTM on OOD datasets by 1.5\%. The ECF algorithm also helped pre-trained models, improving the OOD generalization accuracy by 1.2\% on BERT and by 1.7\% on Roberta. % \uncertain{ok}

\begin{figure}
    \centering
    \includegraphics[height=0.45\columnwidth]{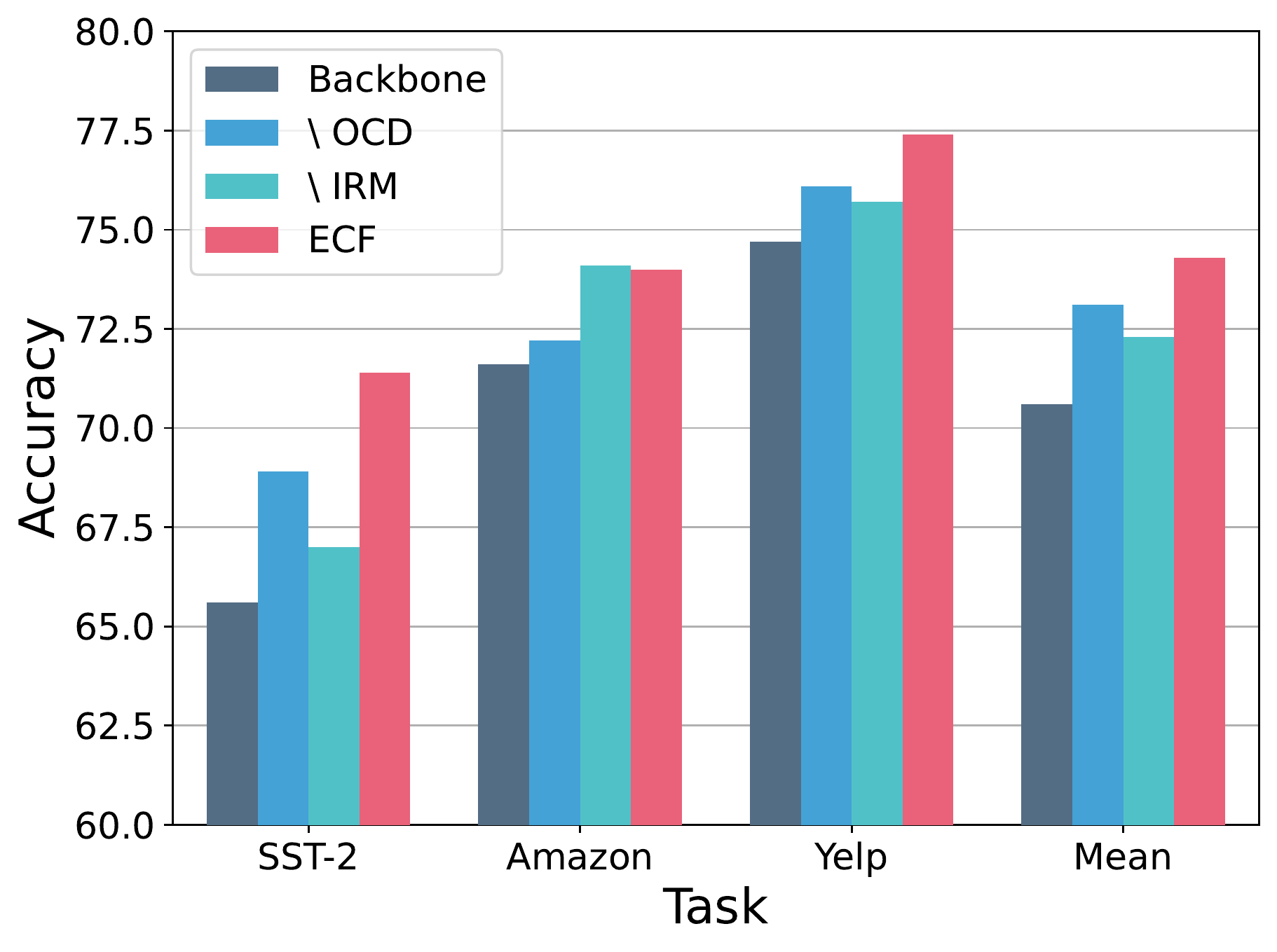}
    % \vspace{-0.4cm}
    \caption{Ablation analysis of two constraints on Sentiment Analysis. $\backslash$ denotes the removing operation. }
    \label{f6-1}
    % \vspace{-0.5cm}
\end{figure}

\subsection{Ablation Study}

We investigated the independent impact of each constraint in our ECF algorithm. We chose BERT as the backbone, and the results are reported in Fig. \ref{f6-1}. When we removed $\mathcal{L}_{IRM}$ and $\mathcal{L}_{OCD}$, the performance decreased significantly. This illustrated that the language models trained directly on CAD did not fully exploit the potential of CAD, and two additional constraints we proposed further unlocked CAD's potential. % \uncertain{ok}

\subsection{Data Efficiency}

\begin{figure}[t]
    \centering
    \subfigure[SA]{
    \includegraphics[height=0.45\columnwidth]{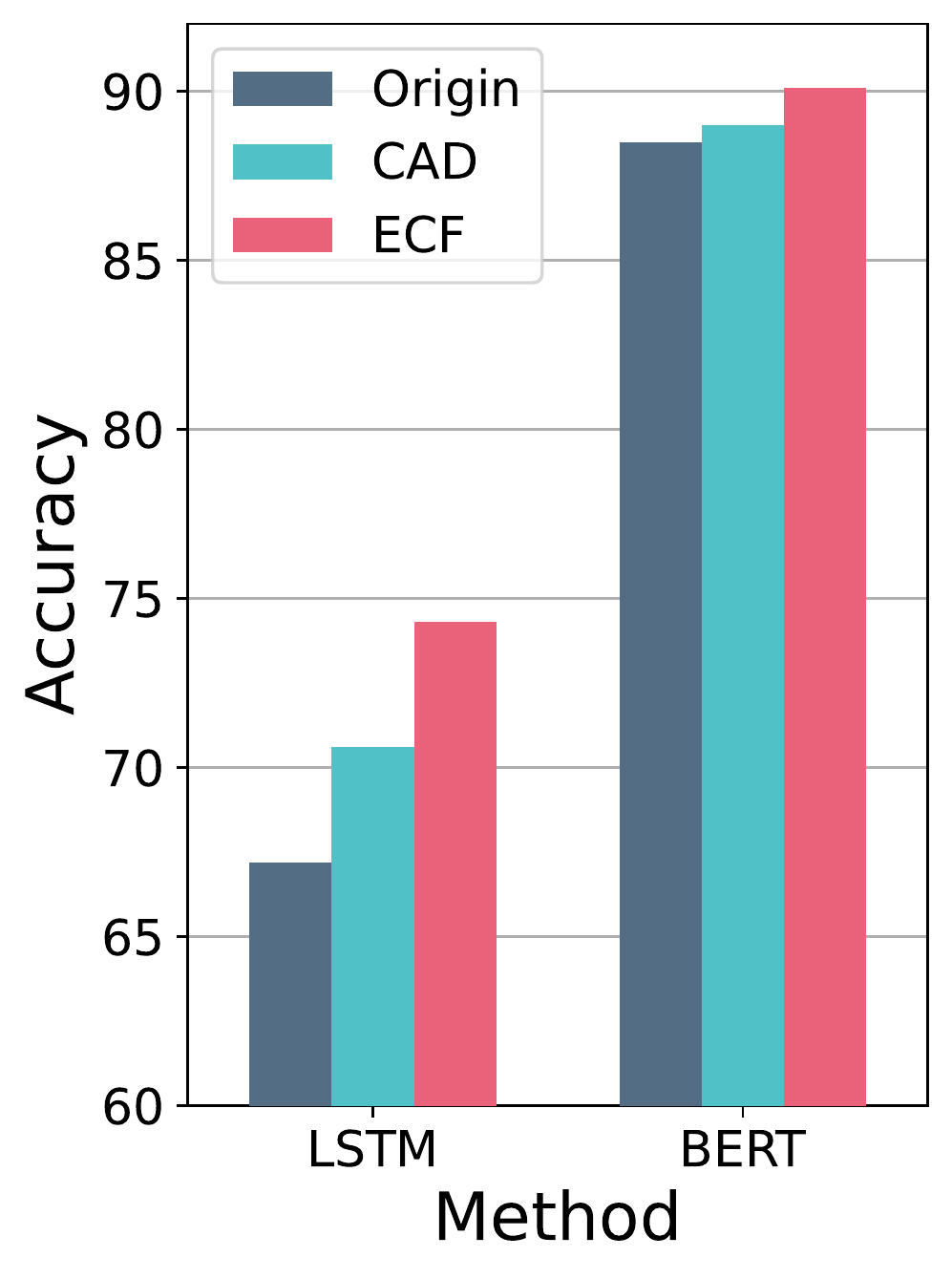} 
    \label{f6-2a}
    }
    \subfigure[NLI]{
    \includegraphics[height=0.45\columnwidth]{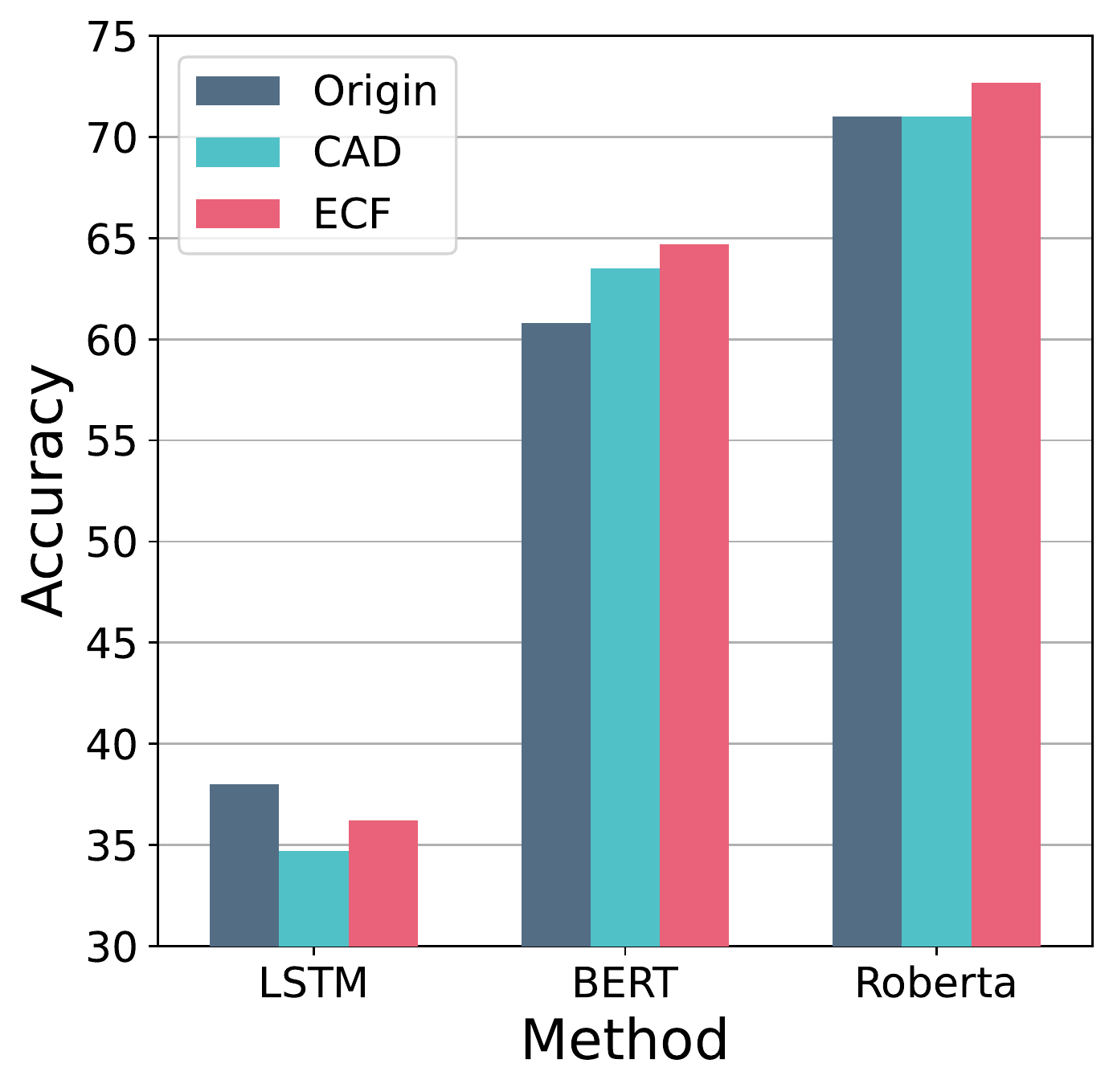} 
    \label{f6-2b}
    }
    % \vspace{-0.4cm}
    \caption{Data efficiency analysis of CAD. }
    \label{f6-2}
    % \vspace{-0.5cm}
\end{figure}

Counterfactual augmentation expanded the size of the seed dataset, which also contributed to OOD generalization. To demonstrate the validity of CAD and our ECF algorithm for language models, we compared the generalization capability of multiple language models trained with the same amount of unaugmented data, as shown in Fig. \ref{f6-2}. The experimental results illustrated that CAD cannot always outperform the same amount of unaugmented data, while our ECF algorithm could steadily improve the generalization capability. % \uncertain{ok}

% \vspace{-0.1cm}

\section{Related Work}

CAD is an emerging technique in NLP field since \cite{Kaushik2020LearningTD}, which aims to help language models extract causal features by changing data distribution. Some studies \cite{Huang2020CounterfactuallyAugmentedST,Khashabi2020MoreBF} pointed out CAD inefficiency in terms of empirical experimental results, and \cite{Joshi2021AnIO} attempted to provide a theoretical explanation for this inefficiency. Previous approaches to improving CAD efficiency fall into two categories: (1) improving the generation quality \cite{Wang2021RobustnessTS,Yang2021ExploringTE} of counterfactual texts. (2) debiasing for specific bias \cite{Balashankar2021CanWI} (e.g., gender, race) in CAD. To the best of my knowledge, our paper is the first attempt to improve the efficiency of CAD by designing additional constraints, which neither change the dataset nor require additional information, and is the most general application scenario. % \uncertain{ok}

\section{Conclusion}

In this paper, we attributed the inefficiency of CAD to \textbf{Myopia Phenomenon} caused by counterfactual augmentation operations, and designed dataset-level and sentence-level constraints based on the structural properties of CAD to help language models to extract more complete causal features and then unlock the potential of CAD. % \uncertain{ok}

% \noindent \textbf{Limitations} \quad There were two main limitations in this paper. First, the assumption about changes to sentence features by the counterfactual augmentation operation was too idealistic, and it was possible that $h_r$ would change as well. In addition, we ignored the interaction between causal components, as shown by the assumption about $h_e$ and $h_u$ in Eq. \ref{e2-1}. This simplification may lead to deviations. % \uncertain{ok}

% \noindent \textbf{Future Work} \quad 

\clearpage
\bibliographystyle{IEEEbib}
\bibliography{reference}

\end{document}